\documentclass[journal, twoside, twocolumn]{IEEEtran}
\ifCLASSOPTIONcompsoc
  \usepackage[nocompress]{cite}

\else
  \usepackage{cite}
\fi
\ifCLASSINFOpdf
\else
\fi
\usepackage{amsmath}
\usepackage{lineno,hyperref}
\usepackage{colortbl}
\usepackage{amsmath,amssymb,amsfonts}
\usepackage{algorithmic}
\usepackage{graphicx}
\usepackage{float}
\usepackage{tikz}
\usepackage{array}
\usepackage{times}
\usepackage{helvet}
\usepackage{subfig}
\usepackage{amssymb}
\usepackage{amsmath}
\usepackage{amssymb}
\usepackage{courier}
\usepackage{multirow}
\usepackage{mathrsfs}
\usepackage{graphicx}
\usepackage{enumitem}
\usepackage{graphicx}
\usepackage{blindtext}
\usepackage{algorithm}
\usepackage{algorithmic}
\usepackage{lineno,hyperref}
\usepackage[marginal]{footmisc}

\usepackage[utf8]{inputenc}
\usepackage[english]{babel}
\usepackage{epstopdf}
\usepackage{array}
\usepackage{multirow}
\usepackage{booktabs}

\usepackage{caption}
\begin{document}

\title{Interpolation-based Contrastive Learning 

for Few-Label Semi-Supervised Learning
}

\author{Xihong~Yang,~Xiaochang~Hu,~Sihang~Zhou,~Xinwang~Liu,~En~Zhu% 

}


\maketitle

\begin{abstract}
Semi-supervised learning (SSL) has long been proved to be an effective technique to construct powerful models with limited labels. In the existing literature, consistency regularization-based methods, which force the perturbed samples to have similar predictions with the original ones have attracted much attention for their promising accuracy. 
However, we observe that, the performance of such methods decreases drastically when the labels get extremely limited, e.g., $2$ or $3$ labels for each category. Our empirical study finds that the main problem lies with the drift of semantic information in the procedure of data augmentation. The problem can be alleviated when enough supervision is provided. However, when little guidance is available, the incorrect regularization would mislead the network and undermine the performance of the algorithm. 
To tackle the problem, we (1) propose an interpolation-based method to construct more reliable positive sample pairs; (2) design a novel contrastive loss to guide the embedding of the learned network to change linearly between samples so as to improve the discriminative capability of the network by enlarging the margin decision boundaries. Since no destructive regularization is introduced, the performance of our proposed algorithm is largely improved. Specifically, the proposed algorithm outperforms the second best algorithm (Comatch) with $5.3\%$ by achieving $88.73\%$ classification accuracy when only two labels are available for each class on the CIFAR-10 dataset. Moreover, we further prove the generality of the proposed method by improving the performance of the existing state-of-the-art algorithms considerably with our proposed strategy.
\end{abstract}


\section{Introduction}\label{sec:introduction}

\IEEEPARstart{I}n recent years, machine learning has developed rapidly and achieved remarkable performance in many fields like, image classification \cite{He_2022_CVPR,2021Robust}, object detection \cite{objectsdection,bekkerman2006target}, semantic segmentation \cite{segmentation,sihang3}, and clustering \cite{liliang,siwei1,sihang_1,siwei2,sihang_2,siwei3,sihang4,siwei4}. Convolutional neural networks (CNNs) have attracted the attention of many researchers. The success of most of these deep neural networks depends heavily on a large number of high-quality labeled datasets\cite{deepnetwork1,2021Robust,he2022towards}.

% \IEEEPARstart{C}onvolutional neural networks (CNNs) have attracted the attention of many researchers for its remarkable performance in applications like, image recognition \cite{2021Robust}, target detection \cite{objectsdection} and semantic segmentation \cite{segmentation}. The success of most of these deep neural networks depends heavily on a large number of high-quality labeled datasets \cite{deepnetwork1}. 
However, collecting labeled data can consume a lot of resources which is un-affordable to countless everyday learning demands in modern society. Therefore, deep learning algorithms which can achieve appropriate performance with tractable supervision have been a hot research spot in recent years. Specifically, deep semi-supervised learning (SSL) algorithms, which seek to improve the performance of deep learning models on datasets with only limited labeled data by leveraging large amounts of unlabeled data, are an important branch in this family. This has led to a plethora of SSL methods designed for various fields \cite{review1_1,review1_2,review1_3,zhangzhao1,review2_1,review2_2,review2_3}.

Among all the deep semi-supervised learning algorithms, consistency regularization based methods treat the original input and its augmented version as positive pairs, which is a form of contrastive learning\cite{constant1,constant2,pimodel,meanteacher,MixMatch,FixMatch,remixmatch,CoMatch}. These consistency regularization-based methods follow a common assumption that ever after data augmentation, the classifier could output the same class probability for an unlabeled sample, which means data augmentation will not change the semantic. The input image should be more similar to its augmented version than other images. Under this assumption, researchers perturb the input samples by conducting data augmentation to generate similar samples of the original data. 
% The classifier will output the same class distribution of the data processed by data augmentation. 
% The loss functions of most SSL are mainly contained of supervised loss and unsupervised loss. 
% The consistency regularization is applied to measure the similarity between output distributions by unsupervised loss. 
% For example, $\pi$ \cite{pimodel} model and MixMatch \cite{MixMatch} use MSE as unsupervised loss, CE is used in FixMatch.

\begin{figure}[!t]
\centering
\includegraphics[width = 1\linewidth]{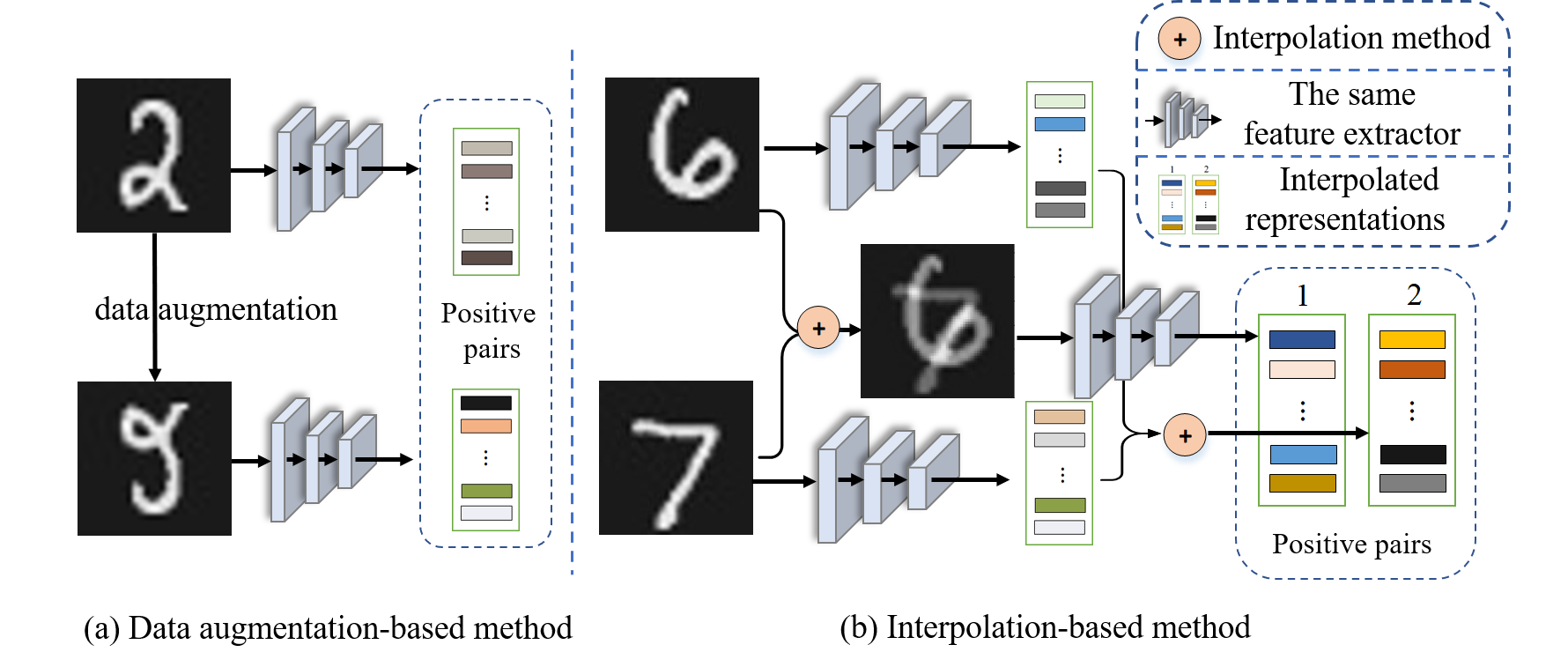}
\caption{Illustration of the positive sample pair construction process. Different from the existing works which construct positive sample pairs with data augmentation, we construct positive sample pairs with interpolation operations. Specifically, given two unlabeled images, the integration of the sample embeddings ’1’ and the embedding of the sample integration ’2’ are acquired as a positive sample pair.}

\label{methodcontrast}  
\end{figure}

The mentioned algorithms have contributed remarkable performance improvement to improve the learning accuracy when only a few labeled data are available. However, we observe that, when the number of labeled data gets extremely small, e.g., $2$ to $3$ labels for each category, the performance of the existing algorithms would drop drastically. For example, to the CIFAR-10 dataset whose scale for training samples is $50,000$ and $10$ categories, the performance of the state-of-the-art algorithm MixMatch \cite{MixMatch} can achieve the top-1 accuracy of $86.47\%$ when $250$ labeled data is available. Nevertheless, the performance of the same algorithm drops to $50.10\%$ when only $30$ labeled samples are available. The similar phenomenon happens to the Mean-Teacher\cite{meanteacher} algorithm whose performance drop by more than a half when the label number decreases from $250$ to $30$. More experimental results can be found in Table \ref{declineacc}.

\begin{table}[]
\caption{Classification accuracy of two state-of-the-art semi-supervised algorithms, i.e., MixMatch\cite{MixMatch} and Mean-Teacher\cite{meanteacher}, on CIFAR-10 dataset with $30, 40, 250, 500$ and $1000$ labels.}
\scalebox{0.9}{
\begin{tabular}{@{}cc|ccccccccc@{}}
\hline
\multirow{2}{*}{\textbf{Method}}       & \multirow{2}{*}{}  & \multirow{2}{*}{\textbf{30}} & \multirow{2}{*}{\textbf{}} & \multirow{2}{*}{\textbf{40}} & \multirow{2}{*}{\textbf{}} & \multirow{2}{*}{\textbf{250}} & \multirow{2}{*}{\textbf{}} & \multirow{2}{*}{\textbf{500}} & \multirow{2}{*}{\textbf{}} & \multirow{2}{*}{\textbf{1000}} \\
                                       &                    &                              &                            &                              &                            &                               &                            &                               &                            &                                \\ \hline
\multirow{2}{*}{\textbf{MixMatch}}     & \multirow{2}{*}{\cite{MixMatch}} & \multirow{2}{*}{50.10}       & \multirow{2}{*}{}          & \multirow{2}{*}{59.08}       & \multirow{2}{*}{}          & \multirow{2}{*}{86.47}        & \multirow{2}{*}{}          & \multirow{2}{*}{89.33}        & \multirow{2}{*}{}          & \multirow{2}{*}{90.79}         \\
                                       &                    &                              &                            &                              &                            &                               &                            &                               &                            &                                \\
\multirow{2}{*}{\textbf{Mean-Teacher}} & \multirow{2}{*}{\cite{meanteacher}} & \multirow{2}{*}{24.51}       & \multirow{2}{*}{}          & \multirow{2}{*}{24.93}       & \multirow{2}{*}{}          & \multirow{2}{*}{52.49}        & \multirow{2}{*}{}          & \multirow{2}{*}{70.15}        & \multirow{2}{*}{}          & \multirow{2}{*}{80.12}         \\
                                       &                    &                              &                            &                              &                            &                               &                            &                               &                            &                                \\ \hline
\end{tabular}}
\label{declineacc}
\end{table}

% visualizing change of augmentation
\begin{figure}[!t]
\centering
\includegraphics[width = 0.8\linewidth]{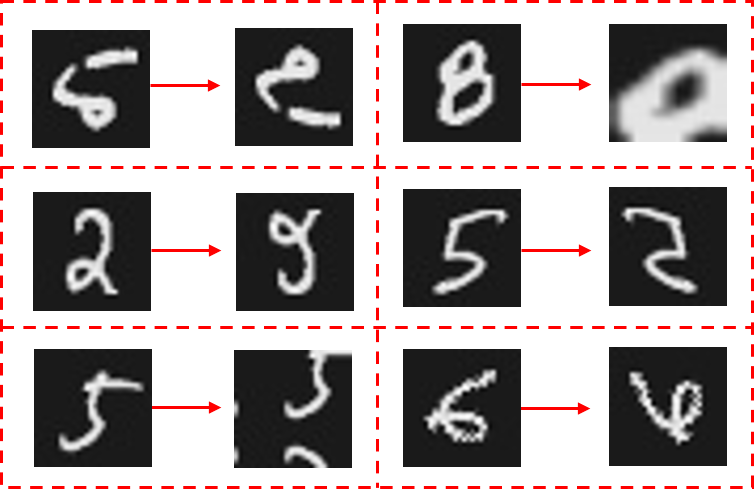}
\caption{Representative examples of semantic information drift caused by inappropriate data augmentation of MINIST samples.}
\label{beforeVerticalFlip}  
\end{figure}

According to our analysis, one of the main reasons that cause large performance decrease lies with the semantic information drift during data augmentation. Taking the samples in the MINIST dataset for example, when the vertical flip is applied to the samples, the labels of "6"s and "9"s, "2"s and "5"s can easily get changed. This would challenge the rationality of the information consistency assumption of existing methods. This problem could be alleviated when relatively abundant label information is available. However, when the label information is extremely lacked, the performance of the corresponding algorithms could decrease a lot.

In this paper, to solve the problem of semantic information drift caused by data augmentation-based positive sample pair construction, we propose a novel interpolation-based positive sample pair construction fashion. Generally, our design roots from the observation that the margin of decision boundaries would get larger if the prediction of the network could change linearly\cite{mixup,ICT}. Under the circumstance of semi-supervised learning, when the label is extremely limited, we seek to improve the discriminative capability of the network by forcing the embedding of the network to change linearly. Specifically, given two unlabeled images, on the one hand, we embed the samples separately into the latent space. On the other hand, we conduct image-level interpolation for an integrated image and do the embedding with the same network. Then, by combining the embedding of the interpolated images with the interpolation of the embeddings, we construct a positive sample pair. In our setting, the negative sample pairs are the embedding pair of different samples. By forcing the positive sample pairs to be close to each other in the latent space and the negative sample pairs to get far away from each other, we enlarge the margin of  decision boundaries, thus improving the performance of the algorithm. To achieve the goal, we further propose a novel contrastive learning-based loss function to guide the network for better learning. We name the resultant algorithm Interpolation Contrastive Learning Semi-Supervised Learning (ICL-SSL).

The main contributions of this paper are listed as follows:
\begin{itemize}
\item We find that semantic information drift is one of the main problems that cause the performance of existing consistency regularization-based semi-supervised algorithms to decrease drastically when extremely limited labeled data is provided. 
\item We propose an interpolation-based positive sample construction method and a novel contrastive loss function to solve the problem and improve the learning accuracy.

\item Our experimental results on the benchmark datasets verify the superior performance of the proposed algorithms against the state-of-the-art algorithms. We also show the generality of our proposed algorithm by enhancing the performance of the existing advanced algorithms steadily with our method.

\end{itemize}

\begin{figure}[!t]
\centering
\includegraphics[width = 0.9\linewidth]{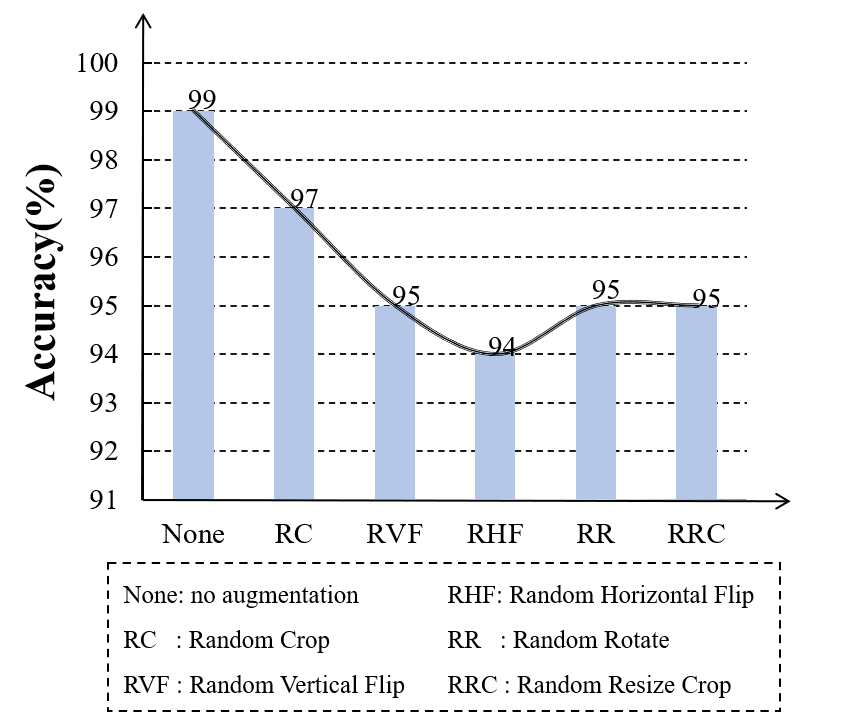} 
\caption{Illustration of classification results with different data augmentations on the MINIST dataset.}
\label{accchange}  
\end{figure}

 

\section{Related Work}\label{relatedWork}
In this section, we first define the main notations and then review several semi-supervised learning (SSL) methods related to our method ICL-SSL.

\subsection{Notations Definition}

Given a dataset $\mathcal{D} = \mathcal{X} \cup \mathcal{U}$, where $\mathcal{X} = \{(x_1,y_1),\cdots, (x_m, y_m)\}$ is an labeled sub-dataset, $\mathcal{U} = u_{m+1}, \cdots, u_{m+n}\}$ is a unlabeled sub-dataset, $n \gg  m$ and $y_m$ is encoded by one-hot, we define a classification model as $p(y|x;\theta)$, which outputs a distribution over class labels $y$ for an input $x$ with parameters $\theta$. For the model $p(y|x;\theta)$, it is concatenated by a encoder network $f(\cdot)$ and a classification head $h(\cdot)$ before softmax function. Meanwhile, after the encoder network $f(\cdot)$, we set a projection head $g(\cdot)$, outputting the normalized low-dimensional representation $z = g(f(\cdot))$. To simplify, $F(\cdot)$ is defined as $g(f(\cdot))$. For more detailed definitions, please refer to Table \ref{notation}.

\begin{table}
\caption{Notation summary}
\scalebox{0.95}{
\begin{tabular}{@{}lll@{}}
\toprule
Notations  & \multicolumn{2}{l}{Meaning}  \\ \midrule
b  & Bathsize &   \\
C   & The numer of classes  &  \\
$\mathcal{X}^B = {(x_i,y_i)}$  & {\color[HTML]{000000} Labeled sub-dataset with  $b$} & {\color[HTML]{FE0000} \textbf{}} \\
$\mathcal{U}^B = \{u_1, \cdots, u_b\}$& {\color[HTML]{000000} Unlabeled sub-dataset matrix  with $b$} & {\color[HTML]{FE0000} \textbf{}} \\
$x, u \in {R}^{C} \times {R}^{H} \times {R}^{W}$ & Input samples&  \\
$y \in \{0, 1\}^C$ & Label with $C$ classes encoded by one-hot &\\
$q \in {R} ^C$  & Predicted category probability distribution &  \\
\textbf{$f(\cdot)$} & The encoder network &  \\
\textbf{$h(\cdot)$} & The classifier  &                                  \\
\textbf{$g(\cdot)$} & The projector head & \\
$z\in R^D =g(f(x)) = F(x)$  & Normalized low-dimensional representation  &  \\  \bottomrule
\end{tabular}}
\label{notation}
\end{table}

\subsection{Contrastive Learning}
Thanks to leveraging unlabeled data for model training, contrastive learning attracts much attention of some researchers and becomes a hot spot recently \cite{SIMCLR,MOCO,BARLOW,SIMSAIM}. It is a widely adopted form of self-supervised learning \cite{contrastlearn1,contrastlearn2,SIMCLR,contrastlearn3,contrastlearn4,DCRN}, which can be used to optimize the task of instance discrimination. Instead of training a classification, contrastive learning is to maximize the similarities of positive pairs and minimize the similarities of negative pairs. It is important to learn the invariance with different views generated by data augmentations. The contrastive learning loss on unlabeled data can be described as follows:

\begin{equation}
{
    -log\frac{exp({F(DA(x_i))}\cdot{F(DA(x_i))}/T)}{\sum_{j=1}^N{ exp(({F(DA(x_i)})\cdot{F(DA(x_j)})/T)}},}
\label{Contrastive_regular}
\end{equation}

where $T$ is a temperature parameter\cite{CoMatch}. {$DA(\cdot)$ denotes the stochastic data augmentation function.} $F(\cdot)$ is the simplified presentation of the encoder network $f(\cdot)$ and the project head $g(\cdot)$. In recent methods, through designing a memory bank, MoCo\cite{MOCO} maintains the consistency of the negative sample pairs. SimCLR\cite{SIMCLR} calculates the pairwise similarity between two similar samples from the images in the same batch, which pushes the negative samples away while pulling the positive samples.  Consistency regularization can be interpreted as a special form of contrastive learning, in which only positive samples are included. 

\subsection{Consistency Regularization}

Consistency regularization utilizes the assumption that the classifier should output the same prediction for the unlabeled data even after it is augmented. Data augmentation is a frequent regularization technique in semi-supervised learning. Through various data augmentation methods, consistency regularization generates a copy of the sample regarded as a similar sample to the original data. In the simplest form, prior work\cite{priorwork1} adds the following consistency regularization loss on unlabeled samples:

\begin{equation}
{
||p(y|DA(x);\theta)-p(y|DA(x);\theta)||_2^2,}
\label{Consitency regular}
\end{equation}
where {$DA(\cdot)$ is a stochastic data augmentation.} With the use of an exponential moving average (EMA) model, Mean-Teacher\cite{meanteacher} replaces one of the terms in Eq.\ref{Consitency regular}, which provides a more stable target. To maximally alter the output class distribution, Virtual Adversarial Training (VAT)\cite{vat} uses an adversarial transformation in place of $DA(\cdot)$. More recently, a form of consistency regularization is utilized in Mixmatch\cite{MixMatch} by using random horizontal flips and crops for the input samples. Unsupervised data augmentation (UDA)\cite{UDA}, ReMixMatch\cite{remixmatch} and FixMatch\cite{FixMatch} have been proposed with the use of weak and strong data augmentations. Generally speaking, through a weakly-augmented unlabeled sample, they generate a pseudo label and enforce consistency against the strongly-augmented version of the same input. 
The above consistency regularization models are based on data augmentation to generate positive samples. Although promising performance has been achieved, we observe that the discriminative capability of previews methods is limited since they would suffer from the semantic information drift issue. Therefore, the constructed samples are no longer similar. Instead of carefully designing data augmentations to utilize consistency regularization, we use an interpolation-based method to obtain positive pairs, which will avoid the semantic information drift caused by data augmentations.

\begin{figure*}[]
\centering
\includegraphics[width=0.8\linewidth]{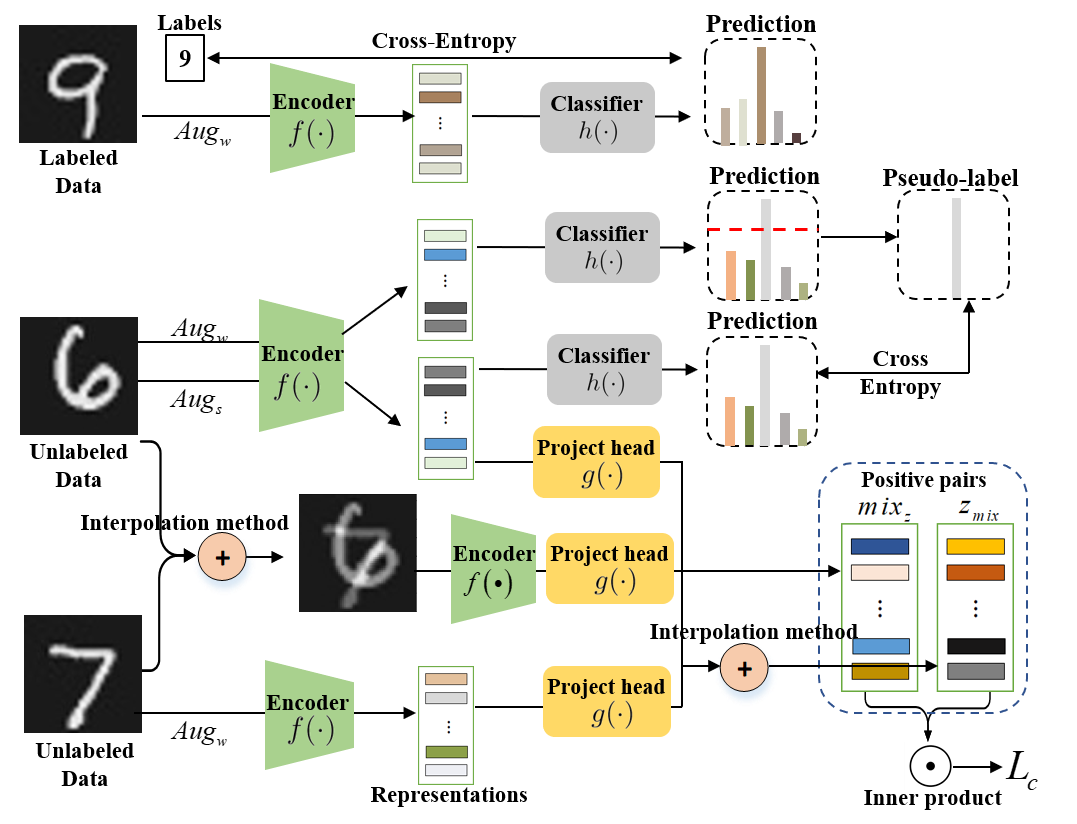} 
\caption{Illustration of Interpolation-based Contrastive Learning Semi-Supervised Learning (ICL-SSL) mechanism. {Following the definition in \cite{FixMatch}, Augw denotes the weak augmentation of the original input image. Augs denotes strong augmentation.} Specifically, given two unlabeled images $u_i$ and $u_j$, we firstly embed the samples separately into the latent space. Then, we conduct image-level interpolation for an integrated image and do the embedding with the same network. $z_{mix}$ and $mix_z$ are the positive embeddings pair constructed by ICL-SSL. By combining these two positive embeddings, we design a novel contrastive loss $L_c$ and force these two embeddings to change linearly between samples, improving the discriminative capability of the network. Therefore, our network would be guided to learn more discriminative embeddings with the interpolation-based method.
}
\label{ICLSSL}  
\end{figure*}

\subsection{The Interpolation-based Method}

Mixup\cite{mixup} is an effective data augmentation strategy for image classification in computer vision\cite{lucasmixed,augmix,guo2020,guo2019}. It linearly interpolates the input samples and their labels on the input data and label spaces.  
% considers a pair of samples $(x_i,y_i)$ and $(x_j.y_j)$. 

\begin{equation}\label{mixup_}
  \begin{aligned}
    \lambda & \sim Beta(\alpha, \beta), \\
    {\lambda}^{\prime}& = max(\lambda, 1-\lambda), \\
    x^{\prime}&={\lambda}^{\prime}{x_1}+(1-{\lambda}^{\prime}){x_2},\\
    y^{\prime}&={\lambda}^{\prime}{y_1}+(1-{\lambda}^{\prime}){y_2},
  \end{aligned}
\end{equation}
where the $\alpha$ and $\beta$ are the parameter of Beta distribution, $\lambda \in [0,1]$. The interpolations of input samples should lead to interpolations of the associated labels. In this manner, Mixup could extend the training distribution. It is recently achieved state-of-the-art performance through different tasks and network architectures. In \cite{toko}, the interpolations are performed in the input space. In order to improve model performance, \cite{berthelot} is proposed to measure the realism of latent space interpolations in unsupervised learning. \cite{ICT} performs the interpolation between input and pseudo-labels. Although the above methods are  verified to be effective, they will still change the construction method of consistency regularized positive sample pairs. Therefore, how to solve the semantic information drift in consistency regularization is an open question. Different from the above approaches, we propose an interpolation-based method term ICL-SSL to construct positive sample pairs. Without using data augmentation to construct positive sample pairs, ICL-SSL is performed between the input samples and the representations, thus avoiding semantic information drift. 

\section{Method}

In this section, we introduce our proposed semi-supervised learning method. Firstly, we will explore the reason for the performance degradation under few labels via some experiments on MINIST dataset. Through the exploratory experiment, we analyze that the semantic information of the input samples will be drifted after some inappropriate data augmentations, thus limiting the performance. After that, to address this issue, we introduce an interpolation-based method ICL-SSL under few labels to construct more reliable positive sample pairs. Finally, we will detail the designed contrastive loss of ICL-SSL.

\subsection{Semantic information drift}

Although promising performances have been achieved by the existing algorithms, we observe that when the number of labeled data gets extremely small, e.g. 2 to 3 labels for each category, the performance of the existing algorithms would decrease drastically. The detailed observation is shown in Table. \ref{declineacc}. Therefore, we conduct experiments to explore the reason to cause the performance dropping. 

Consistency regularization is an essential piece for many state-of-the-art semi-supervised learning methods \cite{MixMatch,remixmatch,FixMatch,CoMatch}. A common assumption of consistency regularization is that the classifier should output the same class probability of an unlabeled sample even if it is augmented. 

In several SSL methods\cite{MixMatch,FixMatch,remixmatch}, when training data is not enough for generalization, data augmentation is a technique to apply consistency regularization. MixMatch\cite{MixMatch} processes the input samples through random horizontal and random crops. The weak data augmentation method uses horizontal flips and vertical flips to process unlabeled samples in FixMatch\cite{FixMatch}. 

Through experiments shown in Fig. \ref{beforeVerticalFlip}, we find that some data augmentations will change the semantic information about the input samples, leading to a decrease in the semantic similarity of the constructed samples damaging the SSL training. We visualize the result of data augmentation. It can be found that the semantic information of the input samples has been changed. Fig. \ref{beforeVerticalFlip} shows that under one data augmentation (random vertical flip), the semantic information of "7"s and "2"s, "6"s and "9"s, "2"s and "5"s can easily get changed. As a result, the quality of the constructed positive samples decreases or the construction fails, which in turn affects the performance of the model. To further verify the effect of data augmentation, we implement experiments on the MINIST dataset.

\begin{figure*}[!ht]
\vspace{20pt}
\begin{center}
{
\centering
\subfloat[CIFAR-10]{{\includegraphics[width=0.22\textwidth]{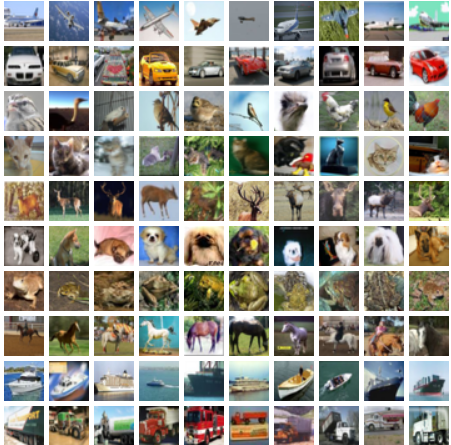}} \label{figure_3a}}\hspace{2mm}
\subfloat[MINIST]{{\includegraphics[width=0.22\textwidth]{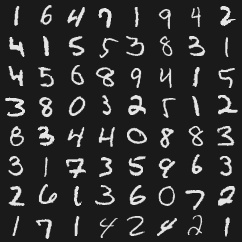}} \label{figure_3c}}\hspace{2mm}
\subfloat[SVHN]{{\includegraphics[width=0.22\textwidth]{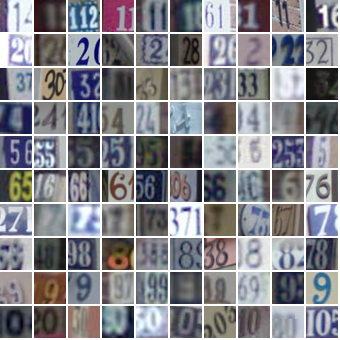}} \label{figure_3d}}\hspace{2mm}
\subfloat[CIFAR-100]{{\includegraphics[width=0.22\textwidth]{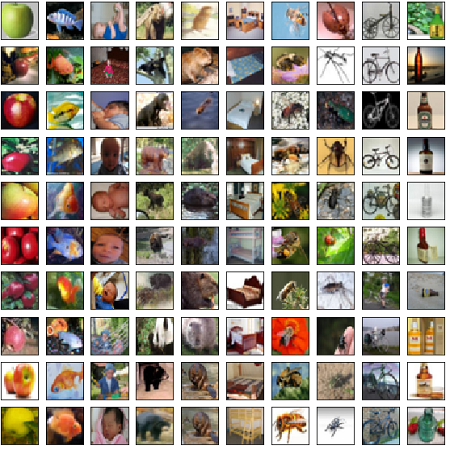}} \label{figure_4d}}

\vspace{3 pt}
\caption{ Illustration of the CIFAR-10, MINIST, SVHN, CIFAR-100 dataset.}
\label{alldataset}
}
\end{center}
\end{figure*}

As shown in Fig. \ref{alldataset}(b), MINIST is a dataset composed of handwritten numbers, which is commonly used in deep learning research. MINIST consists of 60000 training data and 10000 test data. Aiming to reduce the influence of irrelevant factors (e.g. complex structure of training model) to the performance, we explore the semantic information drift problem caused by data augmentation with two-layer MLPs.

From the empirical analysis, we observe that the accuracy is decreased by 5.0\% after the random horizontal flip argumentation on MINIST. As a consequence, after random vertical flips, the accuracy decreases by 4.0$\%$ shown in Fig. \ref{accchange}. Additionally, we also explore rotation, random re-cropping and random cropping, the result shows that those data augmentations will also limit the performance of the model.

The experiment on MINIST can illustrate that during SSL training, some inappropriate data augmentations will change the semantic information of the input samples. Therefore, the semantic correlation of positive sample pairs will be destroyed by inappropriate data augmentations. When the label information is lacking, the incorrect regularization caused by data augmentation would mislead the network and limit the algorithm performance.

% ALGORITHM
\begin{algorithm}[tb]
\small
\caption{Interpolation-based Contrastive Learning Semi-Supervised Learning(ICL-SSL)}
\label{ALGORITHM}
\textbf{Input}: Labeled data $X={(x_1,y_1),(x_2,y_2),....,(x_b,y_b)}$, unlabeled data $U=(u_1,u_2,...,u_b)$, Beta distribution parameter $\alpha$ for feature interpolation, Batch size b, Epoch number e
\begin{algorithmic}[1]
\WHILE{$e$ $\textless$ Epoch} 
\FOR{$i=1$ to $b$}
\STATE{${y_b} = p(y|x)$;}
\STATE{${q_i} = p(y|{u_i})$;}
\STATE{${z_i} = g(f({u_i}))$;}
\ENDFOR

\FOR{$i$ $\in$ {1, . . . , b} and j  $\in$ {1, . . . , b}}
\STATE{$\lambda = Beta(\alpha, \alpha)$};
\STATE{${u_{mix}} = \lambda * {u_i} + (1 - \lambda ) * {u_j};$}
\STATE{${z_{mix}} = g(f({u_{mix}}))$;}
\STATE{${mix_z} = \lambda * {z_i} + (1 - \lambda) * {z_j}$;}
% \STATE{${s_{i,j}} = ({z_{mix}})^T({mix_z}) / ||{z_{mix}}|| \cdot ||{mix_z}||$;}
\ENDFOR
\ENDWHILE

\STATE{Calculate classification loss via Eq.\ref{supervisedloss}, \ref{unsupervisedlOSS} and \ref{contrastiveloss}}

\end{algorithmic}
\end{algorithm}

\subsection{ICL-SSL}

To solve the semantic information drift problem, we proposed a novel interpolation contrastive learning Semi-supervised learning method termed ICL-SSL. Specifically, ICL-SSL does not change the semantic information during the positive pair construction of consistency regularization. In the following, we first obtain the low-dimensional representation $z$ of the unlabeled sample. Then, we describe the interpolation-based positive sample pairs construction method and loss function in detail.

In our ICL-SSL method, the representations are extracted by encoder network $f(\cdot)$. Concretely, for any two unlabeled samples $u_i, u_j$ in a batch of unlabeled sub-dataset $\mathcal{U}^B$, we could obtain their normalized representations $z_i,z_j$ with $\ell ^2$-norm:

\begin{equation}
\begin{aligned}
  z_i=F({u_i}),
  z_i=\frac{z_i}{||z_i||_2},\\
  z_j=F({u_j}),
  z_j=\frac{z_j}{||z_j||_2},
\end{aligned}
\end{equation}

where $F(\cdot)$ is defined as $g(f(\cdot))$, a simple form of encoder network $f(\cdot)$ and project head $g(\cdot)$.

After that, we perform interpolation operations on the normalized low-dimensional feature representations $z_i$ and $z_j$.

\begin{equation}
\begin{aligned}
mix_z &= \lambda{z_i}+(1-\lambda){z_j}\\
&=\lambda F({u_i})+(1-\lambda)F({u_j}),  
\end{aligned}
\end{equation}

where $mix_z$ denotes the interpolated representation of $z_i$ and $z_j$, $\lambda$ is generated by Beta distribution. Simultaneously, unlike the above steps, we first perform an interpolation operation in the sample space ($u_i$, $u_j$) and then get the normalized  low-dimensional feature:

\begin{equation}
{z_{mix}} = F(\lambda{u_i}+(1-\lambda){u_j}),
% g(f(\lambda{u_i}+(1-\lambda){u_j}))
\end{equation}

where $z_{mix}$ is the representation of interpolated input data $u_i$ and $u_j$. The constructed positive sample pair can be presented as follows:

\begin{equation}
[{mix_z},{z_{mix}}].
\end{equation}

{The framework of our proposed ICL-SSL is shown in Fig. \ref{ICLSSL}. ICL-SSL is a semantic-agnostic positive sample construction method. Specifically, we generate one positive sample from the features $z_{mix}$ obtained by interpolating two inputs, and the other $mix_z$ from interpolating the two features of the input. By this setting, both of these positive samples contain the original semantic information of each input ($u_i,u_j$). It has demonstrated that the interpolation operation has the effect to push the decision boundaries away from the class boundaries in \cite{mixup, ICT}. In this manner, with the utilization of our ICL-SSL, the margin decision boundaries would get larger, thus improving the discriminative capability of the network under few labels.}

\subsection{Loss function}
The loss of ICL-SSL mainly consists of three parts: the supervised classification loss $L_x$, the unsupervised classification loss $L_u$ and the contrastive loss $L_c$.

In detail, $L_x$ is the supervised classification loss on the labeled data, which is defined as the cross-entropy between the ground-truth labels and the model's predictions:
\begin{equation}
{L_x} = \frac{1}{B}\sum_{b=1}^{B}H(y_b,p(y|x_b)),
\label{supervisedloss}
\end{equation}

where $x_b$ denotes the labeled data in $\mathcal{X}^B$. $p(y|{x_b})$ is the output of the classifier. $H$ is the cross-entropy between the two distributions $y_b$ and $p(y|{x_b})$. 

For the unlabeled data, its pseudo label $\hat{q_b}$ is generated by the classification head $h(\cdot)$ and the $Softmax$ function. The formula can be described as:

\begin{equation}
\hat{q_b} = Softmax(h(f(u_b))).
\end{equation}
% ${\ell ^2}$

$L_u$ is defined as the cross-entropy between the pseudo-labels and the model's predictions. It can be calculated by:
\begin{equation}
L_u = \frac{1}{\mu B}\sum_{b=1}^{\mu B} \ell\left ( max\left ( \hat{q_b}\ \right ) \ge \tau \right )  H\left ( \hat{q_b},p\left ( y|u_b \right )  \right ), 
\label{unsupervisedlOSS}
\end{equation}
where $\hat{q_b}$ is the predicted probability of pseudo labels. $\ell$ is the function to calculate the loss. When the largest class probability is above the threshold $\tau$, the loss will be calculated. Meanwhile, $\mu$ is used to count the number of valid unlabeled samples. $H$ is the cross-entropy between $q_b$ and $p(y|{u_b})$. 

Through the positive sample pairs $[mix_z, z_{mix}]$ constructed by interpolation strategy, the contrastive loss can be computed as:
\begin{equation}
L_c = -log\frac{exp(({mix_z}\cdot{z_{mix}})/ T)}{\sum_{k=1}^B{I_{\lfloor {k\neq i} \rfloor
}exp(({z_i}\cdot{z_j})/ T)}},
\label{contrastiveloss}
\end{equation}

where $T$ is a temperature parameter. Similar to SimCLR\cite{SIMCLR}, we do not sample negative samples explicitly. Instead, we treat the other examples within a minibatch as negative samples. $I_{\lfloor {k\neq i} \rfloor \in {0,1}}$ is an indicator function. When $k=i$, the value of $I$ is set to 1. The similarity between positive is measured by the inner product. This loss is calculated across all positive samples in a batch. The contrastive loss encourages the model to produce similar representations for positive samples and pushes the negative samples away. The relation of the embedding changes linearly due to the proposed positive sample pairs constructed method. By minimizing Eq.\ref{contrastiveloss}, the margin decision boundaries will be enlarged, thus improving the discriminative of the network.

In summary, the loss function of ICL-SSL can be computed by:
\begin{equation}
L={L_x}+ {L_u} + \alpha {L_c},
\label{total_loss}
\end{equation}

where $L_x$ represents the supervised loss and $L_u$ is the unsupervised loss. $\alpha$ is a trade-off hyper-parameter to control the weight of the total loss. The detailed learning procedure of ICL-SSL is shown in Algorithm \ref{ALGORITHM}.

\section{Experiment}

We evaluate the effectiveness of ICL-SSL on several semi-supervised learning benchmarks. We focus on the most challenging label-scare scenario where few labels are available, e.g., $2$ or $3$ labels for each category. At the same time, our ablation study teases apart the contribution of ICL-SSL components. In addition, we further verify the generality of the proposed method by improving the performance of the existing state-of-the-art algorithms considerably with our proposed strategy.

\subsection{Implementation details}

\begin{table}[]
\centering
\caption{Dataset summary}
\scalebox{1.05}{
\begin{tabular}{cccccc}
\hline
\multirow{2}{*}{\textbf{Dataset}}   & \multirow{2}{*}{\textbf{Size}} & \multirow{2}{*}{\textbf{Train Set}} & \multirow{2}{*}{\textbf{Test Set}} & \multirow{2}{*}{\textbf{Class}} & \multirow{2}{*}{\textbf{Type}} \\
                                    &                                &                                     &                                    &                                 &                                \\ \hline
\multirow{2}{*}{\textbf{SVHN}}      & \multirow{2}{*}{32 × 32}       & \multirow{2}{*}{73257}              & \multirow{2}{*}{26032}             & \multirow{2}{*}{10}             & \multirow{2}{*}{image}         \\
                                    &                                &                                     &                                    &                                 &                                \\
\multirow{2}{*}{\textbf{MINIST}}    & \multirow{2}{*}{28× 28}        & \multirow{2}{*}{60000}              & \multirow{2}{*}{10000}             & \multirow{2}{*}{10}             & \multirow{2}{*}{image}         \\
                                    &                                &                                     &                                    &                                 &                                \\
\multirow{2}{*}{\textbf{CIFAR-10}}  & \multirow{2}{*}{32 × 32}       & \multirow{2}{*}{50000}              & \multirow{2}{*}{10000}             & \multirow{2}{*}{10}             & \multirow{2}{*}{image}         \\
                                    &                                &                                     &                                    &                                 &                                \\
\multirow{2}{*}{\textbf{CIFAR-100}} & \multirow{2}{*}{32 × 32}       & \multirow{2}{*}{50000}              & \multirow{2}{*}{10000}             & \multirow{2}{*}{100}            & \multirow{2}{*}{image}         \\
                                    &                                &                                     &                                    &                                 &                                \\ \hline
\end{tabular}}
\label{datadecribe}
\end{table}

\subsubsection{Datasets \& Metric}

The proposed algorithms are experimentally evaluated on SVHN \cite{svhn}, CIFAR-10 \cite{cifar10/100} and CIFAR-100 \cite{cifar10/100} datasets. 
\begin{itemize}

\item The CIFAR-10 dataset consists of 60000 images of size 32 × 32. The training set of CIFAR-10 consists of 50000 images and the test set consists of 10000 images. The dataset includes ten classes, including images of natural objects such as horse, deer, fork, car and aircraft. 

\item {The CIFAR-100 dataset is similar to the CIFAR-10 dataset and contains 60000 images of the size 32 × 32. The 100 classes in the CIFAR-100 are grouped into 20 superclasses. Each class consists of 500 training images and 100 testing images.}

\item The SVHN dataset includes 73257 training data and 26032 test data of size 32 × 32. Besides, each example is a close-up image of house numbers from 0 to 9.

\end{itemize}
Detailed dataset statistics are summarized in Table \ref{datadecribe}. We use the accuracy metric to evaluate the classification performance. 

\subsubsection{Experiment Settings}

{All experiments are implemented with an NVIDIA 1080Ti GPU on PyTorch platform.}
Following SSL evaluation methods, we evaluate our method on standard SSL benchmarks with the "Wide-ResNet-28" model from \cite{realistic}. Compared with other methods, our model focuses on the challenging label-scare scenario e.g., $2$ or $3$ labels for each category. For CIFAR-10 and SVHN datasets, we train them for 300 epochs until convergence, the batch size chosen by us is 64. {Due to the limited computing resources, the batch size of the all comparison experiments on CIFAR-100 dataset is set to 16.} The weight parameter $\alpha$ to control loss is set to 0.5, and the parameter $\mu$ of the batch size for the control of unlabeled data is set to 1. The learning rate is set to 0.03 for CIFAR-10, CIFAR-100 and SVHN. The threshold $\tau$ is set to 0.95. Besides, our network is trained using SGD optimizer. For our proposed method, we adopt the source data of CoMatch\cite{CoMatch}. To alleviate the impact of randomness, we evaluate the models on 5 runs for each number of labeled points with different random seeds. 
% In all experiments, we use the same set of hyper-parameters.

{In Sub-Section ``Transfer to other models'', the algorithms are implemented with an NVIDIA 1080Ti GPU on PyTorch platform with 40, 250, 500, and 1000 labels on CIFAR-10 dataset. Three state-of-the-art algorithms are compared in our transferring experiments, including MixMatch \cite{MixMatch}, Mean-Teacher \cite{meanteacher} and VAT \cite{vat}. For those algorithms, we reproduce results by adopting their source code with the original settings. The code for the compared algorithms can be downloaded from the authors' website: MixMatch \footnote{https://github.com/google-research/mixmatch}, Mean-Teacher \footnote{https://github.com/siit-vtt/semi-supervised-learning-pytorch}, VAT\footnote{https://github.com/lyakaap/VAT-pytorch}. Specifically, the training epoch is set as 300. The learning rate of the optimizer is set as 0.002 for MixMatch, 0.003 for Mean-Teacher, and 0.01 for VAT.}

\subsection{Comparison with the State-of-the-Art Algorithms}

In this section, six state-of-the-art semi-supervised algorithms are compared to verify the effectiveness of ICL-SSL. The information for the compared algorithms is listed as follows:

\vspace{5pt}
(1) \textbf{CoMatch}\cite{CoMatch}: The class probabilities and low-dimensional embeddings are jointly learned in CoMatch. Through imposing a smoothness constraint to the class probabilities, the quality of pseudo labels could be improved. Overall, CoMatch combines the pseudo-based model, the contrast-loss-based model and the graph-based model to improve the model performance in the case of few labels.

\vspace{5pt}
(2) \textbf{FixMatch} \cite{FixMatch}: For the labeled image FixMatch utilize weak-augmentation to generate the pseudo label. Additionally, for the unlabeled image, the pseudo label is obtained by the high-confidence prediction. And then, the network is trained to predict the pseudo label with the strongly augmented version of the same image.

\vspace{5pt}
(3) \textbf{MixMatch} \cite{MixMatch}: MixMatch jointly optimizes two losses: the supervised loss and unsupervised loss. In detail, cross-entropy is chosen for the supervised loss. The unsupervised loss is the mean square error (MSE) between predictions and generated pseudo labels. MixMatch constructs pseudo labels by data augmentation. With the use of the sharpen function $Sharpen(\cdot)$, MixMatch could improve the quality of pseudo labels. In addition, Mixup is added in the training process, which can construct virtual samples through interpolation.

\vspace{5pt}
(4) \textbf{Virtual Adversarial Training(VAT)} \cite{vat}:  VAT is based on data perturbation. It replaces data augmentation with adversarial transformations. The adversarial transformation can lead to a lower classification error.

\vspace{5pt}
(5) {\textbf{$\pi$-model} \cite{pimodel}: For the same image, data augmentation is used to apply consistency regularization. The loss of $\pi$-model contains the supervised loss and the unsupervised loss. Specifically, the supervised loss is defined as the cross-entropy loss, and the unsupervised loss is the unsupervised consistency loss.}

\vspace{5pt}
(6) \textbf{Mean-Teacher}\cite{meanteacher}: Mean-Teacher is a student-teacher-approach for SSL. The teacher model is based on the average weights of a student model in each update step. In Mean-Teacher, the mean square error loss (MSE) is used as its consistency loss between two predictions. Besides, it uses the exponential moving average (EMA) to update, because the EMA is only updated once per epoch, which can control the model update speed.

\subsection{Performance Comparison}

\subsubsection{CIFAR-10}

To demonstrate the superiority of ICL-SSL, we conduct performance comparison experiments for our proposed ICL-SSL and 4 baselines, including Mean-Teacher\cite{meanteacher}, MixMatch\cite{MixMatch}, FixMath\cite{FixMatch} and CoMatch\cite{CoMatch}. For CIFAR-10 dataset, we evaluate the accuracy of above methods with a varying number of labeled data from 20 to 40. The results are reported in Table. \ref{cifar10result}. For fairness, we create 5 runs for each number of labeled points with different random seeds to alleviate the influence of randomness. We can observe that our method ICL-SSL outperforms all other methods by a significant margin, taking the result on only 2 labeled data in each class for example, ICL-SSL could reach an accuracy of 88.73$\%$. For comparison, at 20 labels the second best algorithm (CoMatch\cite{CoMatch}) achieves an accuracy 83.43$\%$, which is 5.30$\%$ lower than ICL-SSL. ICL-SSL can achieve higher accuracy by using fewer labels.

\begin{table}[]
\centering
\caption{Accuracy comparison with other state-of-the-art methods on five different folds, including Mean-Teacher\cite{meanteacher}, MixMatch\cite{MixMatch}, FixMatch\cite{FixMatch} and CoMatch\cite{CoMatch} on CIFAR-10 dataset. The red and blue values indicate the best and the runner-up results.}
\scalebox{0.95}{
\begin{tabular}{@{}cc|cccccc@{}}
\hline
                                        &                     &                                                     &                                           &                                                     &                                           &                                                     &                    \\
\multirow{-2}{*}{\textbf{Method}}       & \multirow{-2}{*}{}  & \multirow{-2}{*}{\textbf{20 labels}}                & \multirow{-2}{*}{\textbf{}}               & \multirow{-2}{*}{\textbf{30 labels}}                & \multirow{-2}{*}{\textbf{}}               & \multirow{-2}{*}{\textbf{40 labels}}                & \multirow{-2}{*}{} \\ \hline
                                        &                     &                                                     &                                           &                                                     &                                           &                                                     &                    \\
\multirow{-2}{*}{\textbf{Mean-Teacher}} & \multirow{-2}{*}{\cite{meanteacher}} & \multirow{-2}{*}{21.79±0.57}                        & \multirow{-2}{*}{}                        & \multirow{-2}{*}{24.51±0.35}                        & \multirow{-2}{*}{}                        & \multirow{-2}{*}{24.93±0.62}                        & \multirow{-2}{*}{} \\
                                        &                     &                                                     &                                           &                                                     &                                           &                                                     &                    \\
\multirow{-2}{*}{\textbf{MixMatch}}     & \multirow{-2}{*}{\cite{MixMatch}} & \multirow{-2}{*}{38.51±8.48}                        & \multirow{-2}{*}{}                        & \multirow{-2}{*}{50.10±5.81}                        & \multirow{-2}{*}{}                        & \multirow{-2}{*}{59.08±3.04}                        & \multirow{-2}{*}{} \\
                                        &                     &                                                     &                                           &                                                     &                                           &                                                     &                    \\
\multirow{-2}{*}{\textbf{FixMatch}}     & \multirow{-2}{*}{\cite{FixMatch}} & \multirow{-2}{*}{72.63±5.37}                        & \multirow{-2}{*}{}                        & \multirow{-2}{*}{86.65±3.56}                        & \multirow{-2}{*}{}                        & \multirow{-2}{*}{89.69±4.58}                        & \multirow{-2}{*}{} \\
                                        &                     & {\color[HTML]{0000FF} }                             & {\color[HTML]{0000FF} }                   & {\color[HTML]{0000FF} }                             & {\color[HTML]{0000FF} }                   & {\color[HTML]{0000FF} }                             &                    \\
\multirow{-2}{*}{\textbf{CoMatch}}      & \multirow{-2}{*}{\cite{CoMatch}} & \multirow{-2}{*}{{\color[HTML]{0000FF} 83.43±9.20}} & \multirow{-2}{*}{{\color[HTML]{0000FF} }} & \multirow{-2}{*}{{\color[HTML]{0000FF} 88.68±3.79}} & \multirow{-2}{*}{{\color[HTML]{0000FF} }} & \multirow{-2}{*}{{\color[HTML]{0000FF} 90.14±2.86}} & \multirow{-2}{*}{} \\
                                        &                     & {\color[HTML]{FF0000} }                             & {\color[HTML]{FF0000} }                   & {\color[HTML]{FF0000} }                             & {\color[HTML]{FF0000} }                   & {\color[HTML]{FF0000} }                             &                    \\
\multirow{-2}{*}{\textbf{ICL-SSL}}         & \multirow{-2}{*}{\textbf{Ours}} & \multirow{-2}{*}{{\color[HTML]{FF0000} 88.73±5.69}} & \multirow{-2}{*}{{\color[HTML]{FF0000} }} & \multirow{-2}{*}{{\color[HTML]{FF0000} 90.30±3.10}} & \multirow{-2}{*}{{\color[HTML]{FF0000} }} & \multirow{-2}{*}{{\color[HTML]{FF0000} 91.78±2.23}} & \multirow{-2}{*}{} \\ \hline
\end{tabular}}
\label{cifar10result}
\end{table}

\begin{table*}[h]
\centering
\caption{{ Accuracy comparison with $\pi$ model \cite{pimodel}, Mean-Teacher \cite{meanteacher}, MixMatch \cite{MixMatch}, and FixMatch\cite{FixMatch} on CIFAR-100 and SVHN dataset. The red and blue values indicate the best and the runner-up results. The average and std values of the five fold cross validation are reported.}}
\scalebox{1.05}{
\begin{tabular}{cc|ccccccc|ccccccc}
\hline
                                        &                     & \multicolumn{7}{c|}{}                                                                                                                                                                                                                                                                                                                               & \multicolumn{7}{c}{}                                                                                                                                                                                                                                                                                                              \\
                                        &                     & \multicolumn{7}{c|}{\multirow{-2}{*}{\textbf{CIFAR 100}}}                                                                                                                                                                                                                                                                                           & \multicolumn{7}{c}{\multirow{-2}{*}{\textbf{SVHN}}}                                                                                                                                                                                                                                                                               \\ \cline{3-16} 
                                        &                     &                    &                                                     &                                                    &                                                     &                                                    &                                                     &                                                    &                                                    &                                                     &                                           &                                                     &                                           &                                                     &                    \\
\multirow{-4}{*}{\textbf{Method}}       & \multirow{-4}{*}{}  & \multirow{-2}{*}{} & \multirow{-2}{*}{\textbf{200 labels}}               & \multirow{-2}{*}{}                                 & \multirow{-2}{*}{\textbf{400 labels}}               & \multirow{-2}{*}{}                                 & \multirow{-2}{*}{\textbf{800 labels}}               & \multirow{-2}{*}{}                                 & \multirow{-2}{*}{}                                 & \multirow{-2}{*}{\textbf{250 labels}}               & \multirow{-2}{*}{}                        & \multirow{-2}{*}{\textbf{500 labels}}               & \multirow{-2}{*}{}                        & \multirow{-2}{*}{\textbf{1000 labels}}              & \multirow{-2}{*}{} \\ \hline
                                        &                     &                    &                                                     &                                                    &                                                     &                                                    &                                                     &                                                    &                                                    &                                                     &                                           &                                                     &                                           &                                                     &                    \\
\multirow{-2}{*}{\textbf{$\pi$ Model}}     & \multirow{-2}{*}{\cite{pimodel}} & \multirow{-2}{*}{} & \multirow{-2}{*}{8.53±0.25}                         & \multirow{-2}{*}{}                                 & \multirow{-2}{*}{11.67±0.37}                        & \multirow{-2}{*}{}                                 & \multirow{-2}{*}{17.64±1.06}                        & \multirow{-2}{*}{}                                 & \multirow{-2}{*}{}                                 & \multirow{-2}{*}{42.66±0.91}                        & \multirow{-2}{*}{}                        & \multirow{-2}{*}{53.33±1.39}                        & \multirow{-2}{*}{}                        & \multirow{-2}{*}{65.90±0.03}                        & \multirow{-2}{*}{} \\
                                        &                     &                    &                                                     &                                                    &                                                     &                                                    &                                                     &                                                    &                                                    &                                                     &                                           &                                                     &                                           &                                                     &                    \\
\multirow{-2}{*}{\textbf{Mean-Teacher}} & \multirow{-2}{*}{\cite{meanteacher}} & \multirow{-2}{*}{} & \multirow{-2}{*}{7.11±0.06}                         & \multirow{-2}{*}{}                                 & \multirow{-2}{*}{11.54±0.28}                        & \multirow{-2}{*}{}                                 & \multirow{-2}{*}{17.82±0.09}                        & \multirow{-2}{*}{}                                 & \multirow{-2}{*}{}                                 & \multirow{-2}{*}{42.70±1.79}                        & \multirow{-2}{*}{}                        & \multirow{-2}{*}{55.71±0.53}                        & \multirow{-2}{*}{}                        & \multirow{-2}{*}{67.71±1.22}                        & \multirow{-2}{*}{} \\
                                        &                     &                    &                                                     &                                                    &                                                     &                                                    &                                                     &                                                    &                                                    &                                                     &                                           &                                                     &                                           &                                                     &                    \\
\multirow{-2}{*}{\textbf{MixMatch}}     & \multirow{-2}{*}{\cite{MixMatch}} & \multirow{-2}{*}{} & \multirow{-2}{*}{4.55±0.45}                         & \multirow{-2}{*}{}                                 & \multirow{-2}{*}{17.68±0.07}                        & \multirow{-2}{*}{}                                 & \multirow{-2}{*}{26.75±1.13}                        & \multirow{-2}{*}{}                                 & \multirow{-2}{*}{}                                 & \multirow{-2}{*}{92.12±0.06}                        & \multirow{-2}{*}{}                        & \multirow{-2}{*}{94.53±0.43}                        & \multirow{-2}{*}{}                        & \multirow{-2}{*}{95.13±0.04}                        & \multirow{-2}{*}{} \\
                                        &                     &                    & {\color[HTML]{0000FF} }                             &                                                    & {\color[HTML]{0000FF} }                             &                                                    & {\color[HTML]{0000FF} }                             &                                                    &                                                    & {\color[HTML]{0000FF} }                             & {\color[HTML]{0000FF} }                   & {\color[HTML]{0000FF} }                             & {\color[HTML]{0000FF} }                   & {\color[HTML]{0000FF} }                             &                    \\
\multirow{-2}{*}{\textbf{FixMatch}}     & \multirow{-2}{*}{\cite{FixMatch}} & \multirow{-2}{*}{} & \multirow{-2}{*}{{\color[HTML]{0000FF} 9.31±0.08}}  & \multirow{-2}{*}{\textbf{}}                        & \multirow{-2}{*}{{\color[HTML]{0000FF} 24.44±0.35}} & \multirow{-2}{*}{\textbf{}}                        & \multirow{-2}{*}{{\color[HTML]{0000FF} 28.12±0.30}} & \multirow{-2}{*}{\textbf{}}                        & \multirow{-2}{*}{\textbf{}}                        & \multirow{-2}{*}{{\color[HTML]{0000FF} 95.45±0.07}} & \multirow{-2}{*}{{\color[HTML]{0000FF} }} & \multirow{-2}{*}{{\color[HTML]{0000FF} 95.73±0.15}} & \multirow{-2}{*}{{\color[HTML]{0000FF} }} & \multirow{-2}{*}{{\color[HTML]{0000FF} 95.94±0.10}} & \multirow{-2}{*}{} \\
                                        &                     &                    & {\color[HTML]{FF0000} }                             & {\color[HTML]{FF0000} }                            & {\color[HTML]{FF0000} }                             & {\color[HTML]{FF0000} }                            & {\color[HTML]{FF0000} }                             & {\color[HTML]{FF0000} }                            & {\color[HTML]{FF0000} }                            & {\color[HTML]{FF0000} }                             & {\color[HTML]{FF0000} }                   & {\color[HTML]{FF0000} }                             & {\color[HTML]{FF0000} }                   & {\color[HTML]{FF0000} }                             &                    \\
\multirow{-2}{*}{\textbf{ICL-SSL}}         & \multirow{-2}{*}{Ours} & \multirow{-2}{*}{} & \multirow{-2}{*}{{\color[HTML]{FF0000} 14.06±0.52}} & \multirow{-2}{*}{{\color[HTML]{FF0000} \textbf{}}} & \multirow{-2}{*}{{\color[HTML]{FF0000} 26.52±1.20}} & \multirow{-2}{*}{{\color[HTML]{FF0000} \textbf{}}} & \multirow{-2}{*}{{\color[HTML]{FF0000} 33.81±0.63}} & \multirow{-2}{*}{{\color[HTML]{FF0000} \textbf{}}} & \multirow{-2}{*}{{\color[HTML]{FF0000} \textbf{}}} & \multirow{-2}{*}{{\color[HTML]{FF0000} 95.58±0.14}} & \multirow{-2}{*}{{\color[HTML]{FF0000} }} & \multirow{-2}{*}{{\color[HTML]{FF0000} 95.80±0.12}} & \multirow{-2}{*}{{\color[HTML]{FF0000} }} & \multirow{-2}{*}{{\color[HTML]{FF0000} 96.05±0.14}} & \multirow{-2}{*}{} \\ \hline
\end{tabular}}
\label{comparison_experiments}
\end{table*}

\begin{table*}[!ht]
\setlength{\tabcolsep}{4.2mm}
\caption{Applying the interpolation positive sample pair construction mechanism to other state-of-the-art semi-supervised learning algorithms, including MixMatch\cite{MixMatch}, VAT\cite{vat} and Mean-teacher\cite{meanteacher}, on the CIFAR-10 dataset. The blue values represent the results enhanced by our positive sample pair construction mechanism, and the black values are the results of the original model. ’B’ and ’B+O’ represent the baseline and the baseline with our
method, respectively.}
\begin{tabular}{@{}cc|cccccccccccc@{}}
\hline
                                        &                     & \multicolumn{2}{c}{}                                                          &                             & \multicolumn{2}{c}{}                                                          &                             & \multicolumn{2}{c}{}                                                          &                             & \multicolumn{2}{c}{}                                                          &                             \\
                                        & \multirow{-2}{*}{}  & \multicolumn{2}{c}{\multirow{-2}{*}{\textbf{40 labels}}}                      & \multirow{-2}{*}{\textbf{}} & \multicolumn{2}{c}{\multirow{-2}{*}{\textbf{250 labels}}}                     & \multirow{-2}{*}{\textbf{}} & \multicolumn{2}{c}{\multirow{-2}{*}{\textbf{500 labels}}}                     & \multirow{-2}{*}{\textbf{}} & \multicolumn{2}{c}{\multirow{-2}{*}{\textbf{1000 labels}}}                    & \multirow{-2}{*}{\textbf{}} \\ \cline{3-14} 
                                        &                     &                              &                                                &                             &                              &                                                &                             &                              &                                                &                             &                              &                                                &                             \\
\multirow{-4}{*}{\textbf{Method}}       & \multirow{-2}{*}{}  & \multirow{-2}{*}{\textbf{B}} & \multirow{-2}{*}{\textbf{B+O}}                 & \multirow{-2}{*}{\textbf{}} & \multirow{-2}{*}{\textbf{B}} & \multirow{-2}{*}{\textbf{B+O}}                 & \multirow{-2}{*}{\textbf{}} & \multirow{-2}{*}{\textbf{B}} & \multirow{-2}{*}{\textbf{B+O}}                 & \multirow{-2}{*}{\textbf{}} & \multirow{-2}{*}{\textbf{B}} & \multirow{-2}{*}{\textbf{B+O}}                 & \multirow{-2}{*}{\textbf{}} \\ \hline
                                        &                     &                              & {\color[HTML]{0000FF} }                        &                             &                              & {\color[HTML]{0000FF} }                        &                             &                              & {\color[HTML]{0000FF} }                        &                             &                              & {\color[HTML]{0000FF} }                        &                             \\
\multirow{-2}{*}{\textbf{VAT}}          & \multirow{-2}{*}{\cite{vat}} & \multirow{-2}{*}{20.00}      & \multirow{-2}{*}{{\color[HTML]{0000FF} 23.00}} & \multirow{-2}{*}{}          & \multirow{-2}{*}{34.00}      & \multirow{-2}{*}{{\color[HTML]{0000FF} 41.00}} & \multirow{-2}{*}{}          & \multirow{-2}{*}{47.00}      & \multirow{-2}{*}{{\color[HTML]{0000FF} 48.00}} & \multirow{-2}{*}{}          & \multirow{-2}{*}{61.00}      & \multirow{-2}{*}{{\color[HTML]{0000FF} 66.00}} & \multirow{-2}{*}{}          \\
                                        &                     &                              & {\color[HTML]{0000FF} }                        &                             &                              & {\color[HTML]{0000FF} }                        &                             &                              & {\color[HTML]{0000FF} }                        &                             &                              & {\color[HTML]{0000FF} }                        &                             \\
\multirow{-2}{*}{\textbf{MixMatch}}     & \multirow{-2}{*}{\cite{MixMatch}} & \multirow{-2}{*}{57.86}      & \multirow{-2}{*}{{\color[HTML]{0000FF} 61.88}} & \multirow{-2}{*}{}          & \multirow{-2}{*}{86.06}      & \multirow{-2}{*}{{\color[HTML]{0000FF} 86.50}} & \multirow{-2}{*}{}          & \multirow{-2}{*}{87.00}      & \multirow{-2}{*}{{\color[HTML]{0000FF} 89.14}} & \multirow{-2}{*}{}          & \multirow{-2}{*}{90.46}      & \multirow{-2}{*}{{\color[HTML]{0000FF} 91.56}} & \multirow{-2}{*}{}          \\
                                        &                     &                              & {\color[HTML]{0000FF} }                        &                             &                              & {\color[HTML]{0000FF} }                        &                             &                              & {\color[HTML]{0000FF} }                        &                             &                              & {\color[HTML]{0000FF} }                        &                             \\
\multirow{-2}{*}{\textbf{Mean-Teacher}} & \multirow{-2}{*}{\cite{meanteacher}} & \multirow{-2}{*}{24.86}      & \multirow{-2}{*}{{\color[HTML]{0000FF} 26.24}} & \multirow{-2}{*}{}          & \multirow{-2}{*}{42.88}      & \multirow{-2}{*}{{\color[HTML]{0000FF} 45.58}} & \multirow{-2}{*}{}          & \multirow{-2}{*}{53.40}      & \multirow{-2}{*}{{\color[HTML]{0000FF} 54.90}} & \multirow{-2}{*}{}          & \multirow{-2}{*}{66.98}      & \multirow{-2}{*}{{\color[HTML]{0000FF} 68.48}} & \multirow{-2}{*}{}          \\ \hline
\end{tabular}
\label{transfer_result}
\end{table*}

\begin{figure*}[!ht]
\vspace{10pt}
\begin{center}
{
\centering
\subfloat[]{{\includegraphics[width=0.24\textwidth]{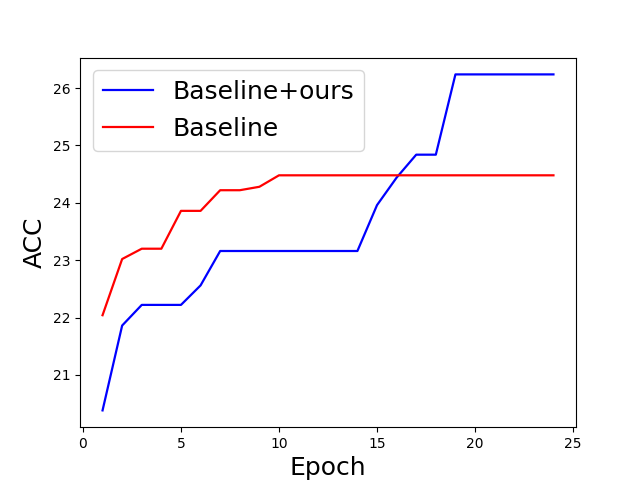}} \label{figure_3a}}%
\subfloat[]{{\includegraphics[width=0.24\textwidth]{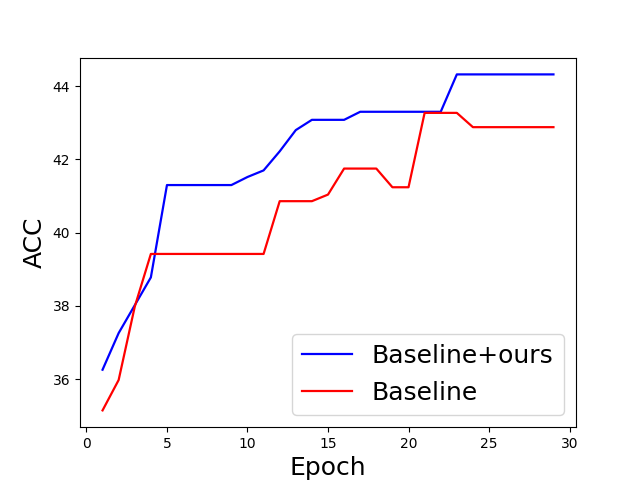}} \label{figure_3c}}
\subfloat[]{{\includegraphics[width=0.24\textwidth]{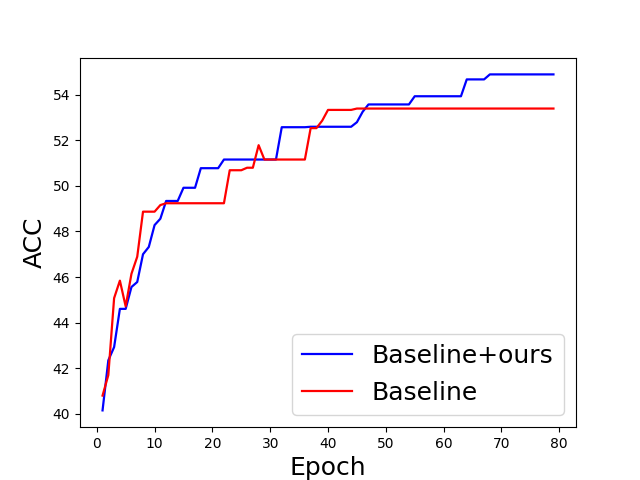}} \label{figure_3d}}%
\subfloat[]{{\includegraphics[width=0.24\textwidth]{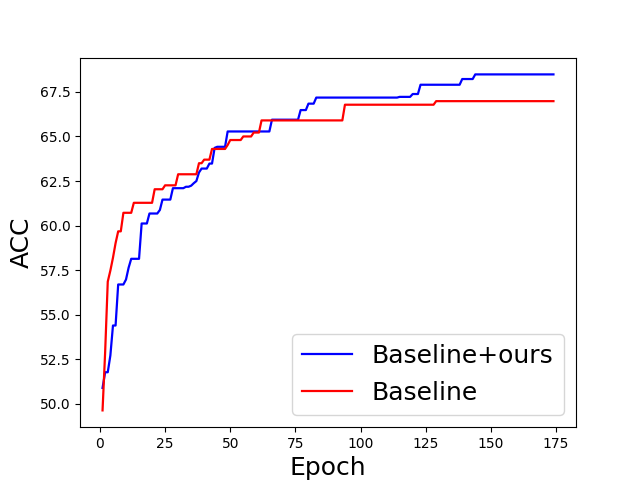}} \label{figure_3f}}
\\
\subfloat[]{{\includegraphics[width=0.24\textwidth]{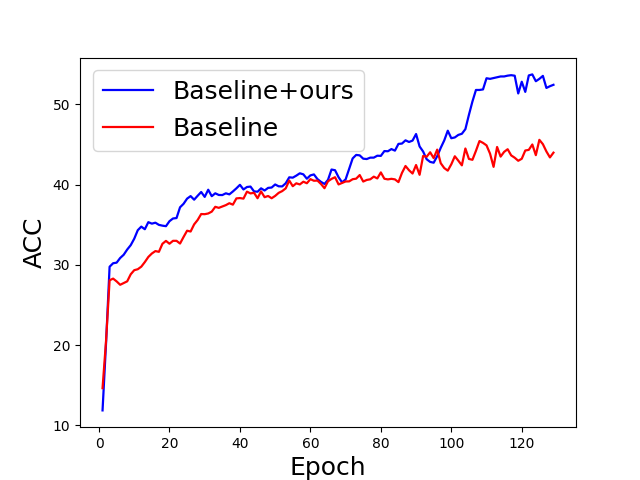}} \label{figure_3a}}%
\subfloat[]{{\includegraphics[width=0.24\textwidth]{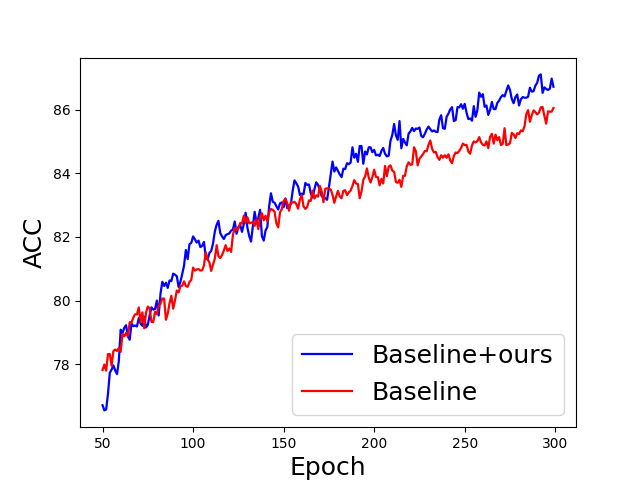}} \label{figure_3c}}
\subfloat[]{{\includegraphics[width=0.24\textwidth]{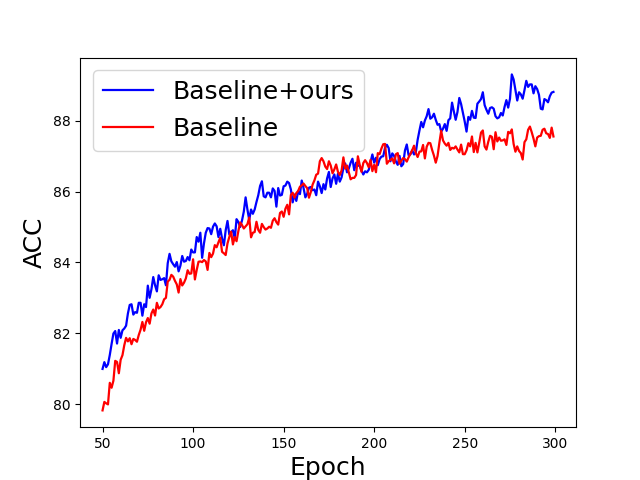}} \label{figure_3d}}%
\subfloat[]{{\includegraphics[width=0.24\textwidth]{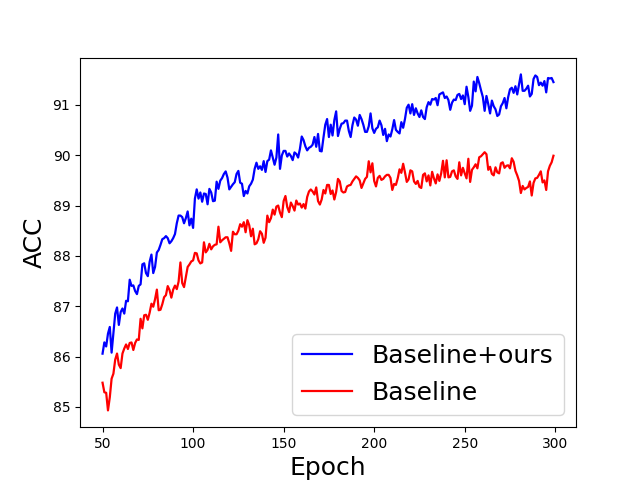}} \label{figure_3f}}
\\

\vspace{3 pt}
\caption{Performance variation when the number of labeled data changes from 40 to 1000 on the CIFAR-10 dataset. (a) - (d) are the classification accuracy of Mean-Teacher \cite{meanteacher} and (e) - (h) are the results of MixMatch \cite{MixMatch}. The blue curve denotes the accuracy enhanced by our positive sample pair construction mechanism, and the red curve represents the accuracy of the original model.}
\label{mt_acc}
}
\end{center}
\end{figure*}

\begin{figure*}[!ht]
\vspace{20pt}
\begin{center}
{
\centering
\subfloat[MixMatch]{{\includegraphics[width=0.3\textwidth]{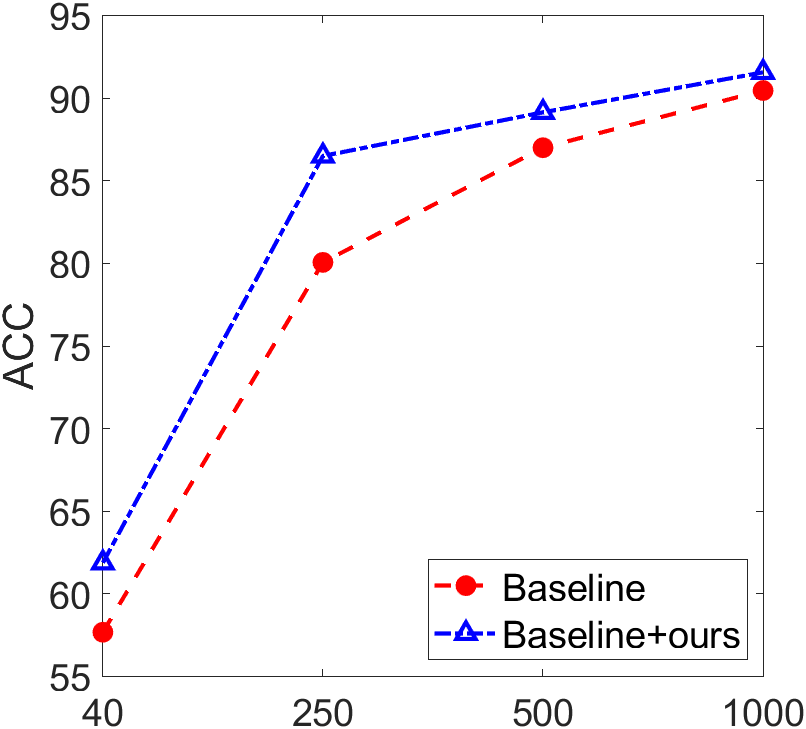}}}\hspace{4mm}
\subfloat[Mean-Teacher]{{\includegraphics[width=0.3\textwidth]{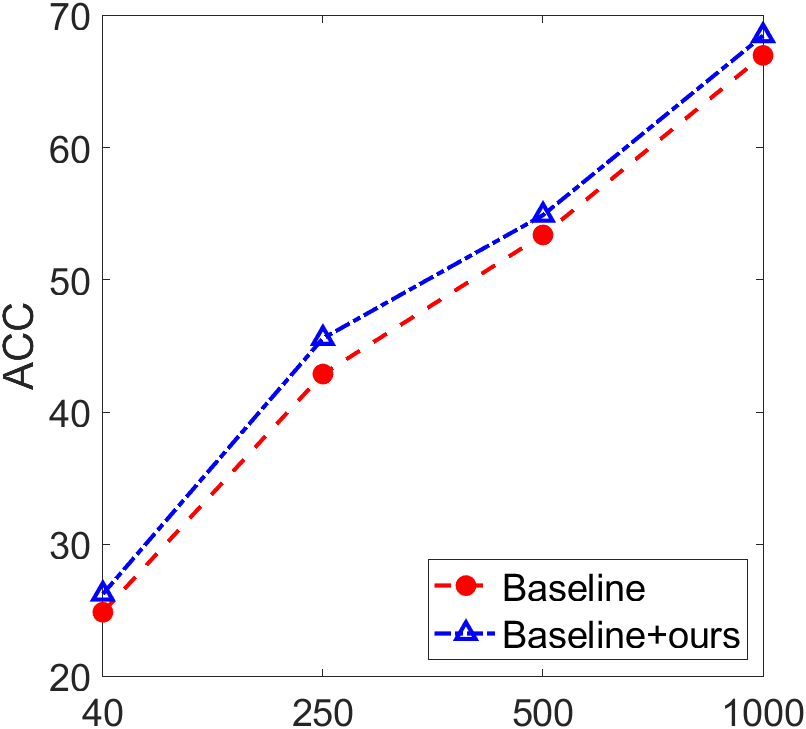}}}\hspace{4mm}
\subfloat[VAT]{{\includegraphics[width=0.3\textwidth]{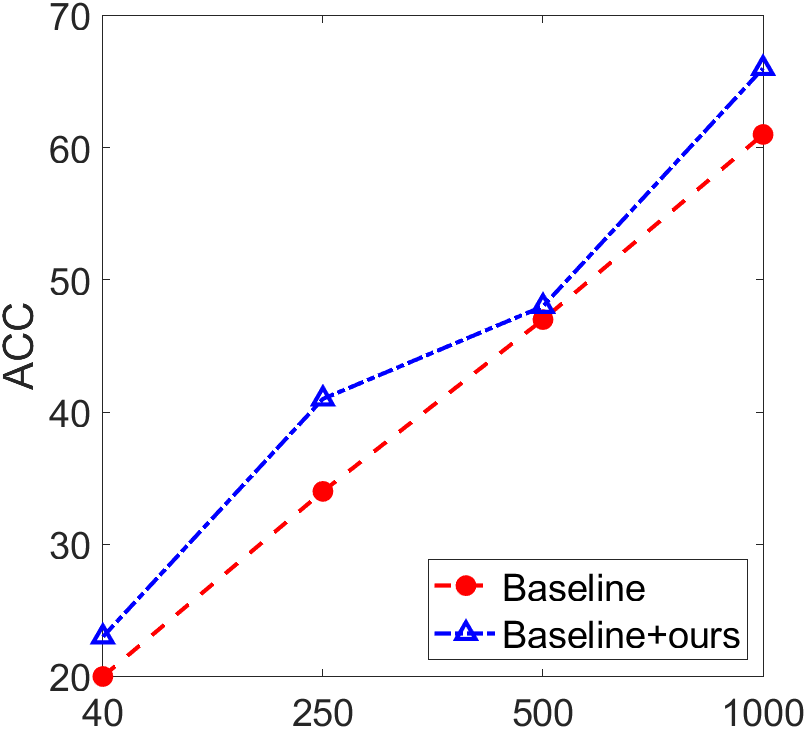}}}
% \subfloat[$\pi$ model]{{\includegraphics[width=0.45\textwidth]{Fig/transfer/pai.png}}}
\\
\vspace{3 pt}
\caption{Performance comparison of different state-of-the-art methods on CIFAR-10 dataset with a varying number of labeled data. 
The blue curve denotes the accuracy enhanced by our positive sample pair construction mechanism, and red curve represents the accuracy of the original model, respectively.}

\label{transfer_curve}
}
\end{center}
\end{figure*}

\subsubsection{SVHN} Moreover, we implement comparison experiments on SVHN dataset. The comparison algorithms contains $\pi$ model \cite{pimodel}, Mean-Teacher \cite{meanteacher}, MixMatch \cite{MixMatch}, FixMatch \cite{FixMatch}. The quantity of labels is 250 to 1000. The results can be seen in Table\ref{comparison_experiments}. With different random seeds, we evaluate the models on 5 runs for each number of labeled data. We could observe that ICL-SSL outperforms all compared methods SVHN with 250, 500, and 1000 labeled data. For example, ICL-SSL exceeds MixMatch by 3.46\% with 250 labels. 

\subsubsection{CIFAR-100}
{To further investigate the effectiveness of our proposed model, we conduct experiments on CIFAR-100 dataset. Table. \ref{comparison_experiments} reports the performance of the four methods with 200, 400, and 1000 labels. From those results, we can observe that, our proposed ICL-SSL could achieve better performance compared with other state-of-the-art algorithms. Taking the result with 200 lables for example, ICL-SSL exceeds FixMatch \cite{FixMatch} by 4.75\%.}

Through the above experiments, our method outperforms all the existing methods in the case of few labels. The reason is that other methods use data augmentation to generate positive sample pairs, easily leading to incorrect regularization. Different from them, our ICL-SSL aims to improve the discriminative capability from two aspects. Firstly, we proposed an interpolation-based method to construct more reliable positive sample pairs, thus alleviating the incorrect regularization. Additionally, we design a contrastive loss to guide the embedding to change linearly in samples, which could enlarge the margin decision boundaries. In summary, we proposed ICL-SSL that could improve the discriminative capability of the network and achieves the top-level performance on CIFRA-10, SVHN, and CIFAR-100 dataset.

\begin{table}[]
\caption{{Ablation comparisons of ICL-SSL mechanism. The results are reported with 20 labels on CIFAR-10 dataset.}}
\scalebox{0.95}{
\begin{tabular}{@{}l|cc@{}}
\hline
\multirow{2}{*}{\textbf{Ablation}}                                       & \multirow{2}{*}{\textbf{20 labels}} & \multirow{2}{*}{\textbf{40 labels}} \\
                                                                         &                                     &                                     \\ \hline
\multirow{2}{*}{\textbf{ICL-SSL}}                                        & \multirow{2}{*}{88.73}              & \multirow{2}{*}{91.78}              \\
                                                                         &                                     &                                     \\
\multirow{2}{*}{\textbf{ICL-SSL without contrastive loss}}               & \multirow{2}{*}{72.63}              & \multirow{2}{*}{89.69}              \\
                                                                         &                                     &                                     \\
\multirow{2}{*}{\textbf{ICL-SSL without the interpolation-based method}} & \multirow{2}{*}{56.91}              & \multirow{2}{*}{70.89}               \\
                                                                         &                                     &                                     \\ \hline
\end{tabular}}
\label{ablation}
\end{table}

\begin{table}[]
\centering
\caption{{Training and inference time comparison on CIFAR-10 dataset with 20 labels.}}
\scalebox{1.05}{
\begin{tabular}{@{}cc|cccc@{}}
    \hline
    {\color[HTML]{000000} }                         &                        &                    &                                              &                             &                                               \\
    \multirow{-2}{*}{{\color[HTML]{000000}\textbf{ Method}}} & \multirow{-2}{*}{}     & \multirow{-2}{*}{} & \multirow{-2}{*}{\textbf{Training Time (s)}} & \multirow{-2}{*}{\textbf{}} & \multirow{-2}{*}{\textbf{Inference Time (s)}} \\ \hline
                                                &                        &                    &                                              &                             &                                               \\
    \multirow{-2}{*}{\textbf{MixMatch}}             & \multirow{-2}{*}{\cite{MixMatch}}    & \multirow{-2}{*}{} & \multirow{-2}{*}{121.50 $\pm$ 0.21}                     & \multirow{-2}{*}{}          & \multirow{-2}{*}{0.60 $\pm$ 0.13}                        \\
                                                &                        &                    &                                              &                             &                                               \\
    \multirow{-2}{*}{\textbf{FixMatch}}             & \multirow{-2}{*}{\cite{FixMatch}}    & \multirow{-2}{*}{} & \multirow{-2}{*}{155.37 $\pm$ 0.10}                     & \multirow{-2}{*}{}          & \multirow{-2}{*}{3.46 $\pm$ 0.07}                        \\
                                                &                        &                    &                                              &                             &                                               \\
    \multirow{-2}{*}{\textbf{CoMatch}}              & \multirow{-2}{*}{\cite{CoMatch}}    & \multirow{-2}{*}{} & \multirow{-2}{*}{571.65 $\pm$ 0.15}                     & \multirow{-2}{*}{}          & \multirow{-2}{*}{1.30 $\pm$ 0.03}                        \\
                                                &                        &                    &                                              &                             &                                               \\
    \multirow{-2}{*}{\textbf{ICL-SLL}}              & \multirow{-2}{*}{Ours} & \multirow{-2}{*}{} & \multirow{-2}{*}{193.81 $\pm$ 0.14}                     & \multirow{-2}{*}{}          & \multirow{-2}{*}{1.08 $\pm$ 0.62}                        \\ \hline
    \end{tabular}}
\label{time_comparison}
\end{table}

\subsection{Time Cost}\label{time}

{As shown in Table \ref{time_comparison}, we compare the training and the inference time of ICL-SSL and other state-of-the-art algorithms, including MixMatch \cite{MixMatch}, FixMatch \cite{FixMatch}, and CoMatch \cite{CoMatch}.
    The results are the average training time for 300 epochs with 20 labels on CIFAR-10 dataset. We observe that the training and the inference time of ICL-SLL are 193.81 seconds and 1.08 seconds, respectively. From the Table.\ref{time_comparison} we find that the computational efficiency of the proposed algorithm is comparable to the MixMatch and FixMatch and is much faster than that of CoMatch.}

\subsection{Ablation Study}
In this section, we implement extensive ablation studies to examine the effect of different components in ICL-SSL. Due to the number of experiments in our ablation study, we perform the study with 20 and 40 labels split from CIFAR-10 dataset. The parameter settings are kept the same with comparison experiments, and the results are shown in Table. \ref{ablation}.

\vspace{6pt}
 \noindent{\textbf{Effective of contrastive loss}}

To further investigate the superiority of the proposed contrastive loss, we experimentally compare our method. Here, we denote the FixMatch\cite{FixMatch} as the baseline. With the experimental results, in the case of few labels, the model performance achieves better performance than that of baselines. Taking the result on CIFAR-10 with 20 labels for example, the accuracy exceeds the baseline by 16.1$\%$ performance increment. From the empirical analysis, it benefits from the contrastive loss to guide the embedding of the network to change linearly between samples to improve the discriminative capability of the network.

\vspace{6pt}
\noindent{\textbf{Effective of interpolation-based positive samples construction method}}.

Additionally, we verify the effectiveness of the interpolation-based positive samples construction method. As shown in Table\ref{ablation}, we can observe that the accuracy would decrease from 88.73$\%$ to 56.91$\%$. The above experiments demonstrate the effectiveness of the interpolation-based positive samples construction method.

\vspace{6pt}
\subsection{Sensitivity Analysis}
% \noindent{\textbf{Hyper-parameter Analysis of $\alpha$}}

\begin{figure}
\centering
\includegraphics[width=0.95\linewidth]{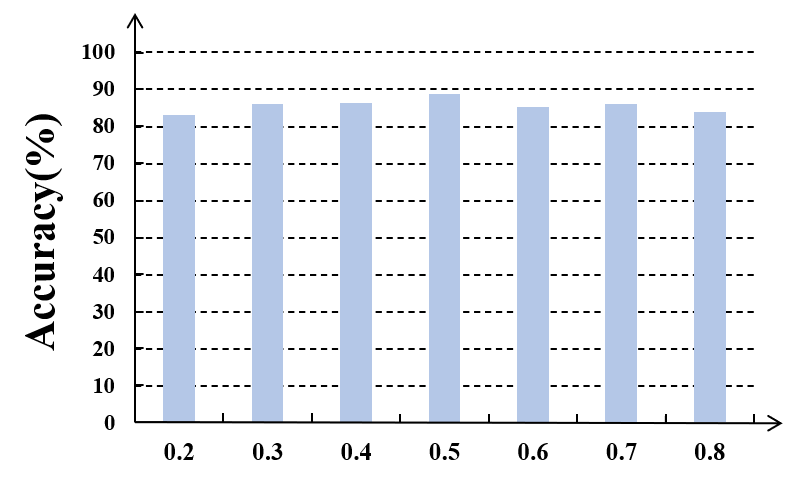}
\caption{{Sensitivity analysis of the hyper-parameter $\alpha$ on CIFAR-10 dataset with 20 labels.}}
\label{loss_weight}
\end{figure}

Further, we investigate the effect of hyper-parameters $\alpha$. {As shown in Fig. \ref{loss_weight}, we observe that the classification accuracy will not fluctuate greatly when the $\alpha$ is varying. This demonstrates that our model ICL-SSL is insensitive to the variation of the hyper-parameter $\alpha$.}

% We observe that when the weight parameter $\alpha$ decreases, the accuracy generally maintains it up to slight variation, indicating that our ICL-SSL is insensitive to the variation of the hyper-parameter $\alpha$. 

\subsection{Transferring to other models}\label{S:f}
To verify the generality of our proposed ICL-SSL, we transfer our method to the existing state-of-the-art algorithms. We implement our method into other semi-supervised learning models (MixMatch\cite{MixMatch}, VAT\cite{vat}, Mean-Teacher\cite{meanteacher}). All the experiments are implemented with CIFAR-10 dataset. Experiments are carried out on the number of labeled data from 40, 250, 500 and 1000. Here, we denote the baseline and the baseline with our method ICL-SSL as ``B'' and ``B+O'', respectively. 

From Fig. \ref{transfer_curve}, we have observed as follows: 1) The models could achieve better performance with our method. 2) As shown in Table \ref{transfer_result}, taking the results in MixMatch \cite{MixMatch} for example, our method could improve the classification accuracy by 4.02\% on 40 labeled data and  2.14\% on 500 labels on CIFAR-10 dataset, respectively. In conclusion, the experiment results show that ICL-SSL can improve the model performance in other semi-supervised models. Moreover, in Fig. \ref{mt_acc}, we further show that other state-of-the-art methods could obtain higher accuracy with our proposed strategy during the training process.

\section{Conclusion} 

In this work, we propose an interpolation-based method termed ICL-SSL to construct reliable positive sample pairs, thus alleviating the semantic information drift with extreme labels (e.g., 2 or 3 labels for each class). Specifically, ICL-SSL is a semantic-agnostic method. We interpolate the input images and their representations in image-level and latent space, respectively. Besides, the designed contrastive loss will guide the embeddings changing linearly between samples and thus get a larger margin decision boundary. Benefiting from this mechanism, the discriminative capability of the network can be improved with extreme labels. Extensive experiments demonstrate the effectiveness and generality of our ICL-SSL. {In the future, we will try to extend ICL-SSL to other fields (e.g. graph semi-supervised node classification). Besides, as we analyzed in section \ref{time}, although our proposed algorithm is as efficient as other state-of-the-art contrastive algorithms, its efficiency still needs to be improved to suit even larger scale datasets. Therefore, how to reduce the training time is also a future work direction.}

\ifCLASSOPTIONcompsoc
  \section*{Acknowledgments}
\else

\bibliographystyle{IEEEtran}
\bibliography{myreference}

\end{document}